## Research Article

# Probability-Density-Based Deep Learning Paradigm for the Fuzzy Design of Functional Metastructures

Ying-Tao Luo[iD],[1] Peng-Qi Li,[1] Dong-Ting Li[iD],[2] Yu-Gui Peng,[1,3] Zhi-Guo Geng[iD],[1] Shu-Huan Xie,[2] Yong Li[iD],[2] Andrea Alù[iD],[3] Jie Zhu[iD],[4,5] and Xue-Feng Zhu[iD][1]

[1]School of Physics and Innovative Institute, Huazhong University of Science and Technology, Wuhan 430074, China
[2]Institute of Acoustics, Tongji University, Shanghai 200092, China
[3]Photonics Initiative, Advanced Science Research Center, City University of New York, 85 St. Nicholas Terrace, New York, NY 10031, USA
[4]Department of Mechanical Engineering, Hong Kong Polytechnic University, Hong Kong SAR, China
[5]Hong Kong Polytechnic University Shenzhen Research Institute, Shenzhen 518057, China

Correspondence should be addressed to Yong Li; yongli@tongji.edu.cn, Andrea Alù; aalu@gc.cuny.edu,
Jie Zhu; jiezhu@polyu.edu.hk, and Xue-Feng Zhu; xfzhu@hust.edu.cn

In quantum mechanics, a norm-squared wave function can be interpreted as the probability density that describes the likelihood of a particle to be measured in a given position or momentum. This statistical property is at the core of the fuzzy structure of microcosms. Recently, hybrid neural structures raised intense attention, resulting in various intelligent systems with far-reaching influence. Here, we propose a probability-density-based deep learning paradigm for the fuzzy design of functional metastructures. In contrast to other inverse design methods, our probability-density-based neural network can efficiently evaluate and accurately capture all plausible metastructures in a high-dimensional parameter space. Local maxima in probability density distribution correspond to the most likely candidates to meet the desired performances. We verify this universally adaptive approach in but not limited to acoustics by designing multiple metastructures for each targeted transmission spectrum, with experiments unequivocally demonstrating the effectiveness and generalization of the inverse design.

## 1. Introduction

Metamaterials have been extensively explored in the past decades, due to their intriguing functionalities that do not exist in nature, such as invisibility cloaking [1–3], superresolution focusing [4–6], zero-index-based funneling [7–10], and abrupt phase modulation in deep-subwavelength scales [11, 12]. With the development of advanced micro-nanoprocessing and 3D printing technologies, versatile metamaterial-based devices have been successfully implemented in practice, with great potentials in designable wave controls for miscellaneous important applications, such as the camouflage and illusion [13], superresolution imaging [14], and metasurface-based lensing and hologram [15–17]. Previously, metadevices with unique functionalities were purposefully designed by following some physical and/or mathematical guiding rules, which belong to the paradigm of forward design. For example, in acoustics, we used transformation acoustics and effective medium approaches to design invisibility cloaks and other illusion devices [2, 18]. For acoustic metamaterials with negative and/or zero refractive indices, we tailored structures with locally resonant unit cells by matching the frequencies of monopole and dipole resonances [19]. To design a specific functional acoustic metasurface, we exploited local resonances to engineer phases of the reflected or transmitted sound [20, 21]. By combining the designed model with powerful finite element solvers, we can in principle achieve complex device functions. Recently, there comes out another paradigm of inverse design of metastructures that fulfill the desired functions [22, 23]. In principle, the inverse design is an optimization process, which is a one-fits-all approach with promising application prospects but hitherto has still represented a challenging task, since it involves the modeling of nondeterministically multivalued physical relation and high-dimensional optimization.

Basically, inverse design should not be a trial-and-error procedure, where the trial-and-error burden will lead to an unpredictable result. Sadly, many current inverse design methods still cannot exclude trial-and-error preprocessing or hyperparameter tuning. In addition, the intuitive trial-and-error procedure requires massive time to search for the optimized structural parameters. Especially, as the structure complexity is increased to match more advanced device functionalities, the parameter space has higher dimensions and the optimization workload becomes enormously heavy. Among all inverse design methods, conventional methods like genetic algorithms [24], adjoint methods [25], and boundary element methods [26] are known to be inaccurate. For example, evolutionary algorithm such as genetic algorithm (GA) requires strong a priori knowledge on the selection of initial population and evaluation function. In most cases, a GA model with random a priori knowledge can easily lead to unacceptable local minima, making it another trial-and-error attempt. Methods such as adjoint methods and boundary element methods are only feasible when an exact partial derivative equation exists and is known for the field of designed structure, which is not the case for many inverse design areas. In recent years, deep learning as an accurate, fast, and adaptive algorithm has been thriving in various fields of research [27], as a way to achieve something never fathomed possible, such as outperforming the champion players in the game of GO. Featured with a huge fitting capacity, deep learning is more powerful than conventional optimization methods in terms of efficiency and accuracy, showing impressive success in discovering new chemical reactions [28] and dealing with problems in many-body quantum physics [29] and density functional theory [30], *etc*. In addition, deep learning-based inverse design has attracted rising attention in the field of metamaterials [31]. Previous studies have uncovered that the corresponding relation between metamaterial functionality and structures usually is not a deterministic injection or surjection, but a one-to-many multivalued problem [32]. This claim is in line with our intuition that different structures may have similar properties. Unfortunately, the baseline artificial neural network model is not capable of addressing multivalued problem, as many recent works have pointed out.

One interesting work, for example, proposed tandem neural network (TNN) that pretrains a forward network and then uses the pretrained module to posttrain the final inverse network [33]. This technique can ensure the convergence of loss, but it limits the choice of inverse designs, due to the fact that the TNN makes a concession by reducing the one-to-many relation into a one-to-one relation. Similar approaches like data preprocessing and backward training also only produce a single plausible metastructure for each input functionality. The choice of inverse designs is vital especially in interdisciplinary real-world scenarios. As the complexity of nowadays metamaterials is drastically increasing, some predicted designs may be hard or uneconomic to be built in nature. Many disciplines face the similar problem, such as machine-learning-based drug discovery and protein design. A protein has endless possibilities of folding and sequence control, which may easily make the machine generalize to a prediction that is unable to be tested or synthesized. While a one-to-one modeling representing the traditional inverse design methods may optimize to a structure that is inoperable in real practice, we are interested in enriching our design choices to avoid bad designs. In other interesting works, deep generative models, such as generative adversarial network (GAN) and variational autoencoder (VAE), are leveraged to feasibly model the one-to-many function to solve that problem, but their convergence is unstable and sensitive to the selection of hyperparameters [34–36]. We often need to employ bilevel optimization strategies such as Bayesian optimization (BO) or grid search to perform a for-loop circle to search for the best hyperparameters so as to improve the model accuracy to an acceptable extent, which increases the time cost by orders of magnitude and makes them again trial-and-error procedures, just as in the case of GA [37]. In all, the currently available algorithms are not ample for the fast and accurate design of functional metastructures and many other materials as well.

Above all, the motivation here is to propose an inverse design paradigm to stably and accurately approximate multivalued function that governs the core of inverse design of acoustic metastructures. It is reasonable to efficiently model the quality-factor distribution of metastructures in the global parameter space, where the fuzzy-featured quality factor represents the likelihood of the designed structure to be the on-demand version. In this work, we propose such quantum-inspired probability-density-based deep learning to address the multivalued problems for the inverse design of advanced functional metastructures. The proposed approach is not trial and error, as the deployment of such approach does not require data preprocessing or hyperparameter tuning. We demonstrate this approach in acoustics by retrieving the best fitting metastructures for a targeted sound transmission dispersion over a wide frequency spectrum, corresponding to the local maxima of probability density distribution in the high-dimensional parameter space. With the measurement of sound transmission of the designed structures in a commercial standing wave tube, we prove that the proposed paradigm has excellent performance. Other previously developed techniques can hardly meet the same metrics. For example, the TNN excels in both efficiency and accuracy, but it has only one output solution [38]. The VAE provides multiple solutions but sacrifices either accuracy or time consumption [39, 40].

## 2. Results

*2.1. Probability-Density-Based Deep Learning Architecture.* The proposed probability-density-based deep learning inverse design have two modules that combine deep learning with mixture Gaussian sampling, as shown in Figure 1. In this hybrid architecture, the front end is a neural network that maps a target transmission spectrum to the parameters of individual Gaussian distributions, other than giving the outputs of the metastructures directly. The rear end uses these parameters to construct a mixture Gaussian distribution by linearly superposing individual distributions and then probabilistically samples output solutions for the metastructure design. Here, the amplitude of mixture Gaussian can be deciphered as the probability of plausible metastructures to fulfill the desired functions. The proposal of probability-density-based network (PDN) gets inspiration from the concept of probability density in quantum mechanics, which measures the likelihood of a particle in a given position or momentum, utilizing the principle of maximum likelihood estimation. It is shown that PDN is actually equivalent to several ANNs with each of them fitting a single target (details discussed in Supplementary Materials (available here)).

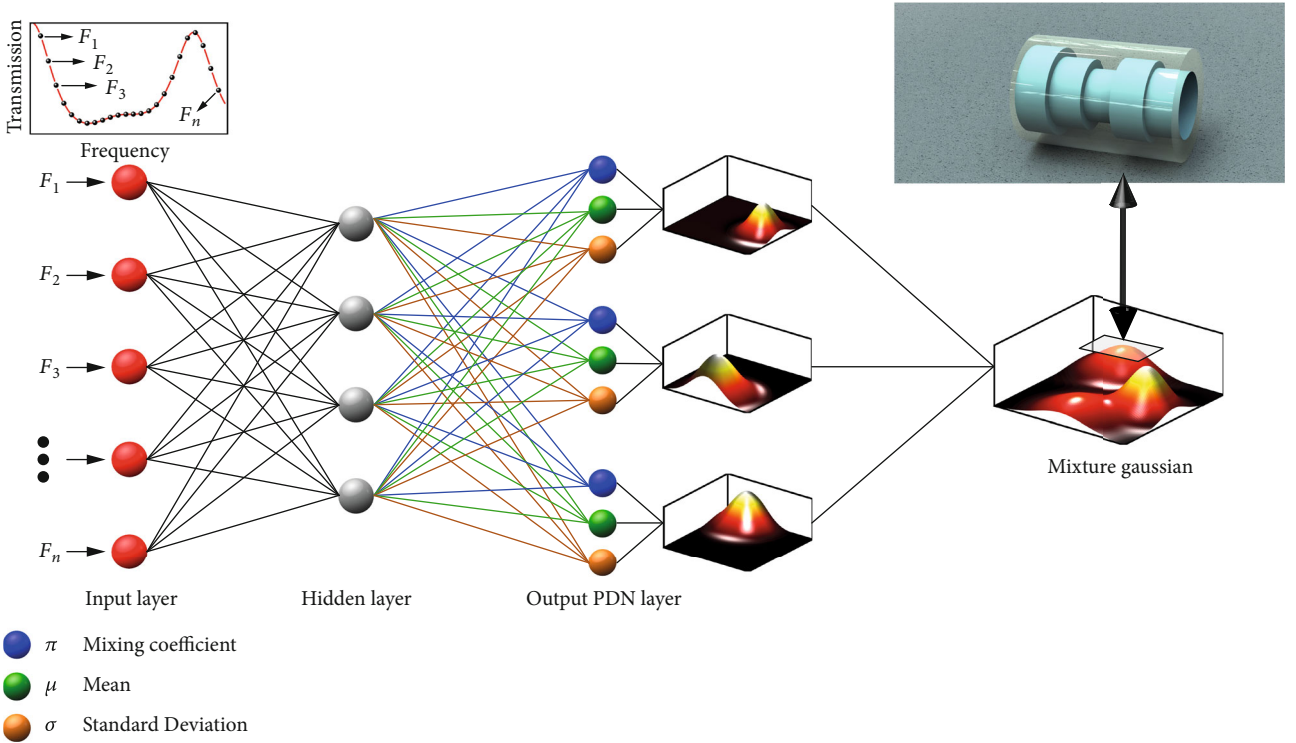

FIGURE 1: Architecture of the proposed neural network. The input of the PDN is the targeted transmission spectrum, while the output is a mixture Gaussian that provides a probabilistic sampling to derive a plausible metastructure with similar transmission. Here, the mixture Gaussian is linearly superposed by individual Gaussians in the output PDN layer that are characterized by the parameters of mixing coefficient $\pi$, mean $\mu$, and deviation $\sigma$. Three output Gaussians are plotted as an example. The local maximum in the mixture Gaussian is mapping to an inversely designed structures with transmission spectra mostly close to the target one.

In Figure 1, the front end is represented by fully connected layers that repeat

$$z^{(i+1)} = \text{ReLU}\left(W^{(i)} \times z^{(i)} + b^{(i)}\right), \quad (1)$$

where $\text{ReLU}(\bullet)$ denotes the rectified linear unit function and $W$ and $b$ are the network parameters, collectively dubbed as $\theta$ in the following for convenience. In Equation (1), the first input is the target transmission spectrum $x$, that is,

$$z^{(0)} = x. \quad (2)$$

For this neural network, the output of the last layer is a mixture Gaussian that can give the estimated approximation of label $y$ and others. Here, the label data $y$ is a group of metastructures that have been tagged for the network training. In our model, the output of the last layer is input back into the parameterization to construct an updated mixture Gaussian distribution, with parameterization expressed into

$$\pi = \text{softmax}\left(W_\pi \times z^{(\text{last})} + b_\pi\right), \quad (3)$$

$$\mu = W_\mu \times z^{(\text{last})} + b_\mu, \quad (4)$$

$$\sigma = \text{Exp}\left(W_\sigma \times z^{(\text{last})} + b_\sigma\right), \quad (5)$$

where $\pi$, $\mu$, and $\sigma$ are the mixing coefficient, the mean, and the standard deviation of the output Gaussian, respectively. In Equation (3), $\text{softmax}(\bullet)$ is an activation function that ensures the summation of $\pi$ to be unitary. In Equation (5), $\text{Exp}(\bullet)$ is an exponent activation function that configures the value of $\sigma$ to be positive. The output mixture Gaussian constructed by the three parameters $\pi$, $\mu$, and $\sigma$ in Equations (3), (4) and (5) provides feasible output solutions for the inverse design of metastructures via probabilistic sampling. Here, the individual Gaussian is denoted by $N(\mu, \sigma)$. Since the output parameters $\pi$, $\mu$, and $\sigma$ are only determined by the input $x$ and the network parameters $\theta$, we can express the generated Gaussian distribution into the form of $Z \sim N(\mu(x, \theta), \sigma(x, \theta))$. The mixture Gaussian, as a sum of individual Gaussian distributions, thus takes the form

$$Z \sim \sum_{i=1}^{m} \pi_i(x, \theta) N(\mu_i(x, \theta), \sigma_i(x, \theta)), \quad \sum_{i=1}^{m} \pi_i(x, \theta) = 1, \quad (6)$$

which can be viewed as a probabilistic sample of model output, giving the probability of plausible metastructures to fulfill the desired functions. In Equation (6), $m$ denotes the number of Gaussians mixed. The PDN that maps the target transmission firstly to a mixture Gaussian and then to the mostly plausible metastructures can be described by conditional probability $p(Z = z \mid X = x, \theta)$, which can be written into

$$p(z \mid x, \theta) = \sum_{i=1}^{m} \pi_i(x, \theta) D(z, \mu_i(x, \theta), \sigma_i(x, \theta)), \quad \sum_{i=1}^{m} \pi_i(x, \theta) = 1,$$
(7)

with

$$D(z, \mu_i, \sigma_i) = \frac{1}{\sigma_i \sqrt{2\pi}} e^{-(z-\mu_i)^2/2\sigma_i^2}. \quad (8)$$

In the training process, we set the optimization objective to maximize the likelihood of metastructure labels in the dataset. We expect that the probabilistic sample has a better chance to get close to the labels. Therefore, the conditional probability of all the labels, $p(y \mid x, \theta)$, must be maximized. The optimization objective in the training can be expressed by the maximum likelihood estimation (MLE)

$$\begin{aligned} \theta^{\text{MLE}} &= \arg\ \max_\theta \log(p(y \mid x, \theta)) \\ &= \arg\ \max_\theta \sum_{i=1}^{n} \log(p(y_i \mid x_i, \theta)), \end{aligned} \quad (9)$$

where $n$ denotes the number of batch-training samples. For any desired transmission spectrum at the input, the PDN will output a mixture Gaussian. Since optimization dictates that the labelled metastructures should have large probabilistic density, several peaks (or local maxima) are likely to appear in the mixture Gaussian, each of which represents a plausible metastructure. The essence of the PDN is that its loss function is not determined by the difference between output and labels, which explains the reason that conventional neural networks always fail in approximating a multifunction. Instead, the PDN learns to produce plausible metastructures by updating the mixture Gaussian, which introduces multiple optimal outputs. In reverse design, because the error is inevitable in the PDN, the confidence in each model prediction is a very important indicator. While conventional neural networks are not capable to rate each prediction's confidence, PDN allows us to evaluate model uncertainty by providing a confidence factor, *i.e.*, the mixture probabilistic density function. As the output of PDN shows the probability of plausible metastructures to fulfill desired functions, we can take local maxima as the best candidates for the functional metastructure design.

*2.2. PDN-Based Reverse Design and Experimental Demonstration.* To give a demonstration, we employ the proposed PDN to inversely design variable cross-sectional metastructures based on the target transmission spectrum in acoustics. As the deep learning is based on a data-driven model, we need to prepare the data before training the network. To collect the label data, we utilize a commercial finite element solver COMSOL Multiphysics 5.3™ that is linked to MATLAB to consider the thermoacoustic effect. The training dataset was collected by uniformly sampling the metastructure parameters, as uniform sampling can avoid bias in the solution space. Specifically, when the metastructure has 5 cylindrical layers with radius for each layer sampled from 8 discrete values, we will end up with a number of training data being $8^5$, that is, 32768, with the sampled radii at each layer to be either 1.8125 mm, 3.625 mm, 5.4375 mm, 7.25 mm, 9.0625 mm, 10.875 mm, 12.6875 mm, or 14.5 mm. We use uniform sampling to avoid bias in the solution space as much as possible, but random sampling is also fine. In this work, the input of the PDN has 250 dimensions, corresponding to transmittances at frequencies from 20 Hz to 5000 Hz with an interval of 20 Hz, while the output has only 5 dimensions of the radii of the five cylindrical layers in metastructures. We also randomly sampled 1000 test data from the continuous range of structural parameters without overlap with training data.

Here, the reverse design of a 5-layer metastructure is a representative case for testing the model's multivalued inverse design capability. We can also apply the same proposed PDN framework for more complicated inverse design, such as advanced acoustic functional metastructures with a large number of layers or structural parameters. This approach can be further extended to other applied disciplines, such as drug synthesis, industrial process optimization, molecular design, and optical device design.

For the scope of the metastructure inverse design, here, we explore two different examples with acoustic experiments to verify the effectiveness of the PDN model. We firstly employed the labelled dataset to train the PDN model and then fixed the weights for inference. In the first example, the target transmission spectrum is featured with a wide bandgap in the frequency range from 1000 Hz to 5000 Hz. With the target spectrum as the input, the PDN model outputted a mixture Gaussian, as shown in Figure 2(a). It needs to be mentioned that the output mixture Gaussian is supposed to be 5 dimensions ($r_1$, $r_2$, $r_3$, $r_4$, and $r_5$), that is, the radii of layers 1-5. Here, to visualize the result, we utilize the technique of principal component analysis [41], a widely used dimension reduction method in data science, to reduce data dimensions from 5 to 2 ($x$ and $y$). We have illustrated the details of this technique in Supplementary Materials. The mixture Gaussian possesses a continuous profile with the amplitude of the confidence of choice for a plausible metastructure, which is also dubbed as the probabilistic density for sampling. With the mixture Gaussian map, we can evaluate the quality factor that characterizes the likelihood of the predicted metastructure to fulfill the input target, even without numerically testing the sample. Since the local maxima in mixture Gaussian correspond to the most confident samples, we chose to directly sample at the peaks in Figure 2(a), as marked by the arrows. In Figures 2(b) and 2(c), the predicted outputs at the locally highest confidence are located at $A_1$ (−0.11, 0.58), $A_2$ (0.89, −0.56), and $A_3$ (−0.76, −0.78) in the reduced 2D space, mapping to 14.29, 12.31, 10.10, 1.89, and 7.85 mm; 13.60, 7.13, 1.90, 11.38, and 12.45 mm; and 11.23, 2.05, 6.95, 11.67, and 13.66 mm in the full 5D space. As shown in Figure 2(c), the results show that, for the three sampled structures, the predicted transmission spectra (blue solid line) are consistent with the target one (red dashed line) and the experimentally measured one (solid spheres).

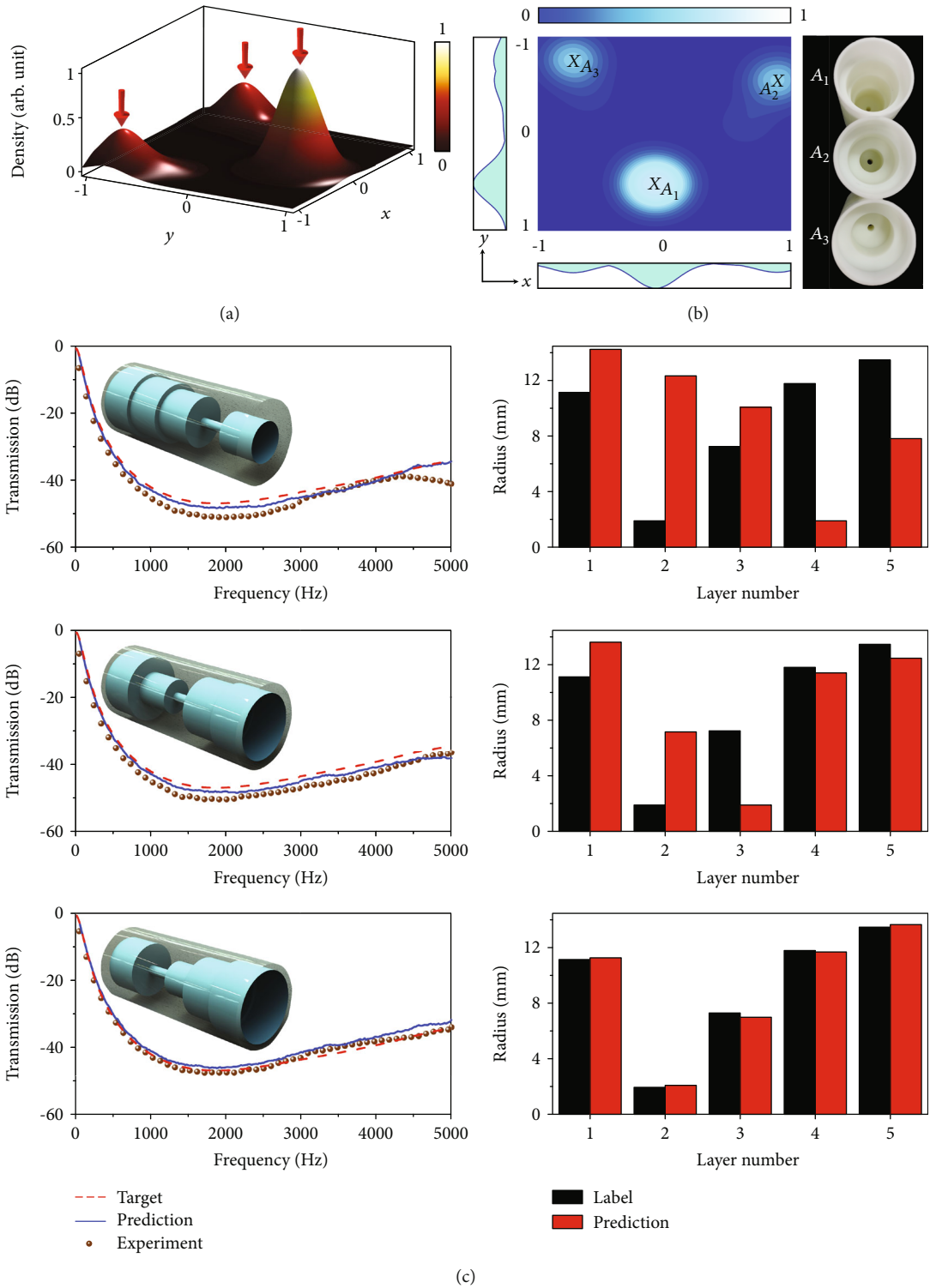

FIGURE 2: Example I of the PDN-based inverse design and experimental demonstration. (a) The output mixture Gaussian for a target transmission spectrum, which is visualized in a 2D plot by reducing the data dimensions from 5 to 2 via the principal component analysis. (b) Exact positions of the local maxima visualized in the contour diagram, where each maximum is mapped to a metastructure. The local maxima and the corresponding metastructures are labelled as $A_1$, $A_2$, and $A_3$, respectively. (c) The comparison among the target transmission spectra, the PDN predicted, and the experimentally measured ones for the three different metastructures $A_1$, $A_2$, and $A_3$ from the top down. Each predicted $A_1$, $A_2$, and $A_3$ structure has different radii per layer, but all get very similar desired transmissions to the target ones. The radii of cylindrical air channels in the labelled structure and the predicted structures are appended on the right side for reference.

In the second example, we introduce one transmission peak at 1000 Hz in the target transmission spectrum. In this case, as shown in Figure 3(a), the PDN outputs a mixed distribution with four local maxima, two of which are very close to each other. In Figures 3(b) and 3(c), the predicted outputs at the locally highest confidence are at $B_1$ (0.58, −0.32), $B_2$ (−0.35, −0.91), $B_3$ (0.58, 0.12), and $B_4$ (−0.28, 0.78) in the reduced 2D space. In the 5D parameter space, the counterparts are at 3.41, 10.94, 9.61, 3.62, and 7.02 mm; 8.84, 3.52, 5.45, 13.71, and 3.55 mm; 3.14, 12.56, 7.62, 3.51, and 8.72 mm; and 7.34, 3.63, 14.35, 3.50, and 6.92 mm, respectively. The similarly structured samples $B_1$ and $B_3$ associated with two closely adjacent peaks in the mixture Gaussian in Figures 3(a) and 3(b) show the advantage of sampling from a continuous probabilistic density distribution, because we can directly and accurately evaluate the confidence of sampling a plausible metastructure from the probabilistic density. In Figure 3(c), a good agreement was experimentally demonstrated among the target transmission spectra, the PDN-predicted ones, and the experimentally measured ones. Both the results presented in Figures 2 and 3 proved the generalization ability of PDN, as the radius of predicted metastructures can land in the interpolated area of uniform sampling points. For example, the predicted $A_2$ structure (13.60, 7.13, 1.90, 11.38, 12.45) mm in Figure 2 has its first layer of radius 13.60 mm in the range of 12.6875–14.5 mm and its fourth layer of radius 11.38 mm in the range of 10.875–12.6875 mm. This is only possible when the PDN can generalize. It is also discussed in the Supplementary Materials and Figs. S4–S5 that transmission of PDN-predicted interpolated structure may not always be alike in the direct average interpolation between two transmissions. This shows that direct interpolation cannot replace the PDN. All these suggest that the PDN is not a simple look-up search in the training dataset, but a complete modeling of inverse relation.

2.3. PDN versus Other Deep Learning Models. In this section, we compare the performance of different deep learning models, which include ANN, GAN, GAN+BO, VAE, VAE+BO, TNN, and PDN, to achieve the task of inversely designing metastructures for the same target. We aim to clarify how well the PDN could compete with other previously reported models. Three indicators are carefully recorded for comparison, which are the mean errors, time consumption, and the output variety during the training or testing process. The mean error represents the gap between the transmission spectra of the predicted structure and the desired acoustic functionality. In each model, we used the same setting for common hyperparameters to ensure the fairness of comparison. For the generative models of GAN and VAE, we arrange two different scenarios. One is to set their unique hyperparameters, *i.e.*, the dimensions of latent spaces and lengths of label condition, without a priori knowledge for simulating the average performance. The other is to employ Bayesian optimization to search for the most suitable hyperparameters in simulations of the best possible performances. Here, we aimed to evaluate the three metrics of performance: output accuracy, inverse design efficiency, and output variety, to make the judgement quantitatively.

In Figure 4, the results clearly demonstrate that our proposed PDN outperforms the other models with balanced qualities in all three aspects. For the mean errors in the training or testing processes, the normalized error values show that only the TNN and PDN have satisfactory accuracies. Even though the TNN leads in accuracy, it has only one output, without variety. For the output variety, the PDN, GAN, and VAE are multivalued. For the time cost, all the models have the same efficiency in case that careful optimizations of hyperparameters are not adopted. The mean errors of those generative models can be further reduced at the trial-and-error expense of time cost, although it is still not satisfactory enough in our task. It might be due to the unbalanced scale between class condition and the latent space of our inverse design. The generative model was originally proposed as an unsupervised learning technique that pushes the generative distribution towards the target distribution, whereas the PDN here is proposed to solely solve the supervised learning of multivalued function fitting. The conditional generative model dealing with finite class labels has proven to be very successful, as a transition to the supervised learning, but there are infinite labels in our case of metastructure design.

As shown in Figure 4, the errors of the GAN and VAE are oscillating, especially for the GAN as the result of convergence towards Nash equilibrium. On the other hand, the training of the PDN is as smooth as the ANN or TNN. It has been explained in Supplementary Materials that the superposed output of the PDN makes it equivalent to a superposition of multiple ANNs, with each ANN fitting only a single label. A baseline ANN model simply with multiple outputs, on the contrary, can only model the cumulative average of all labels, which is meaningless. With an overall consideration of all three aspects, we find that the PDN prevails over the other models in terms of inverse design due to a balanced performance. Moreover, the PDN does not give one certain output or several certain outputs but rates all the plausible structures in the probability that the designed metastructure has the desired functionality. Evaluations of robustness and generalization ability of the proposed PDN are included in

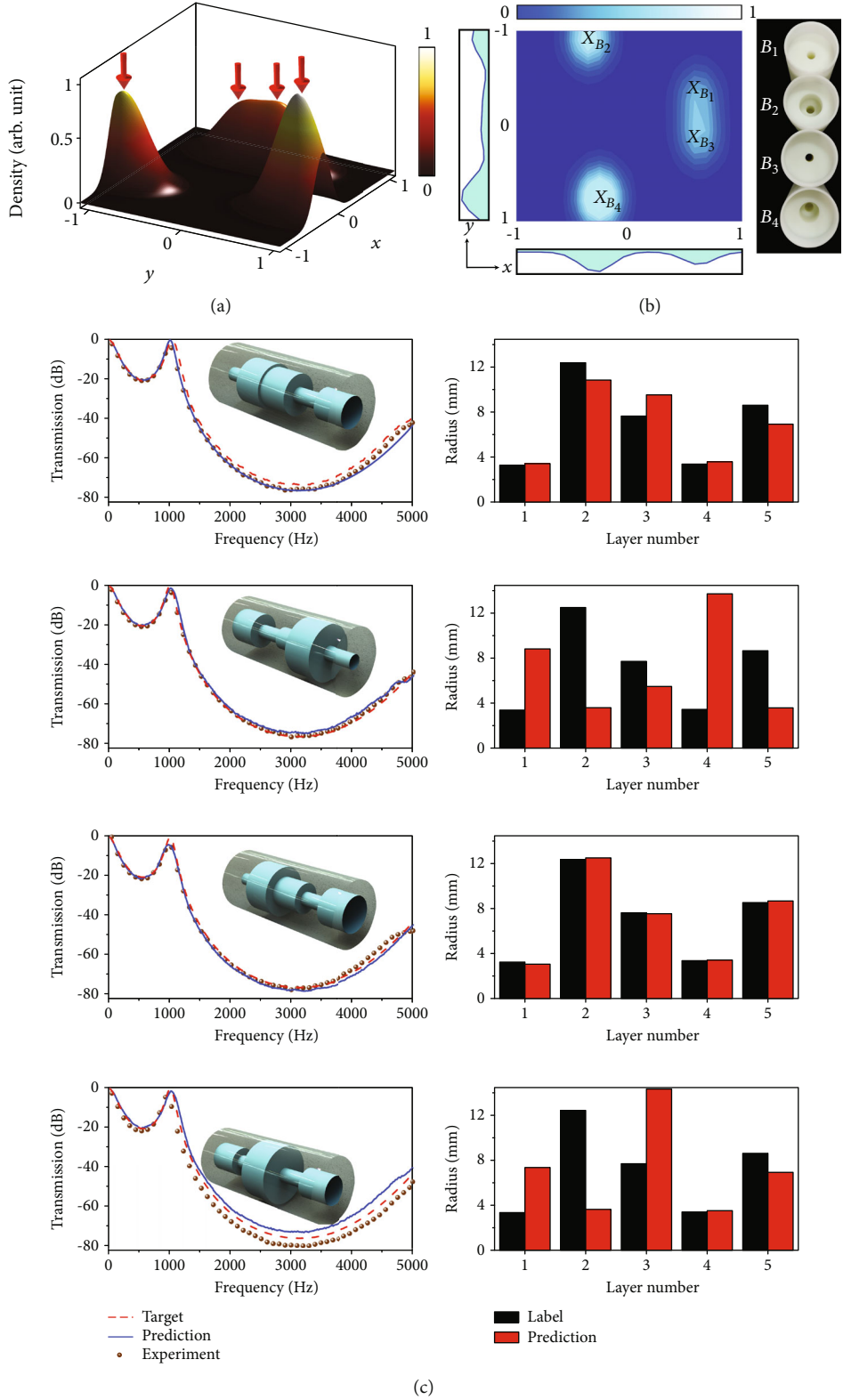

FIGURE 3: Example II of the PDN-based inverse design and experimental demonstration. (a) The output mixture Gaussian for another type of transmission spectrum, which is visualized in a reduced 2D plot. (b) The exact positions of four local maxima visualized by the contour diagram, where the maxima at $B_1$, $B_2$, $B_3$, and $B_4$ correspond to four different metastructures. (c) Comparison among the target transmission spectra, the PDN predicted, and the experimentally measured ones for the metastructures $B_1$, $B_2$, $B_3$, and $B_4$ from the top down. The radii of channels in the labelled structure and the predicted structures are also presented. Each predicted $B_1$, $B_2$, $B_3$, and $B_4$ has different radii, but all get very similar desired transmissions to the target ones.

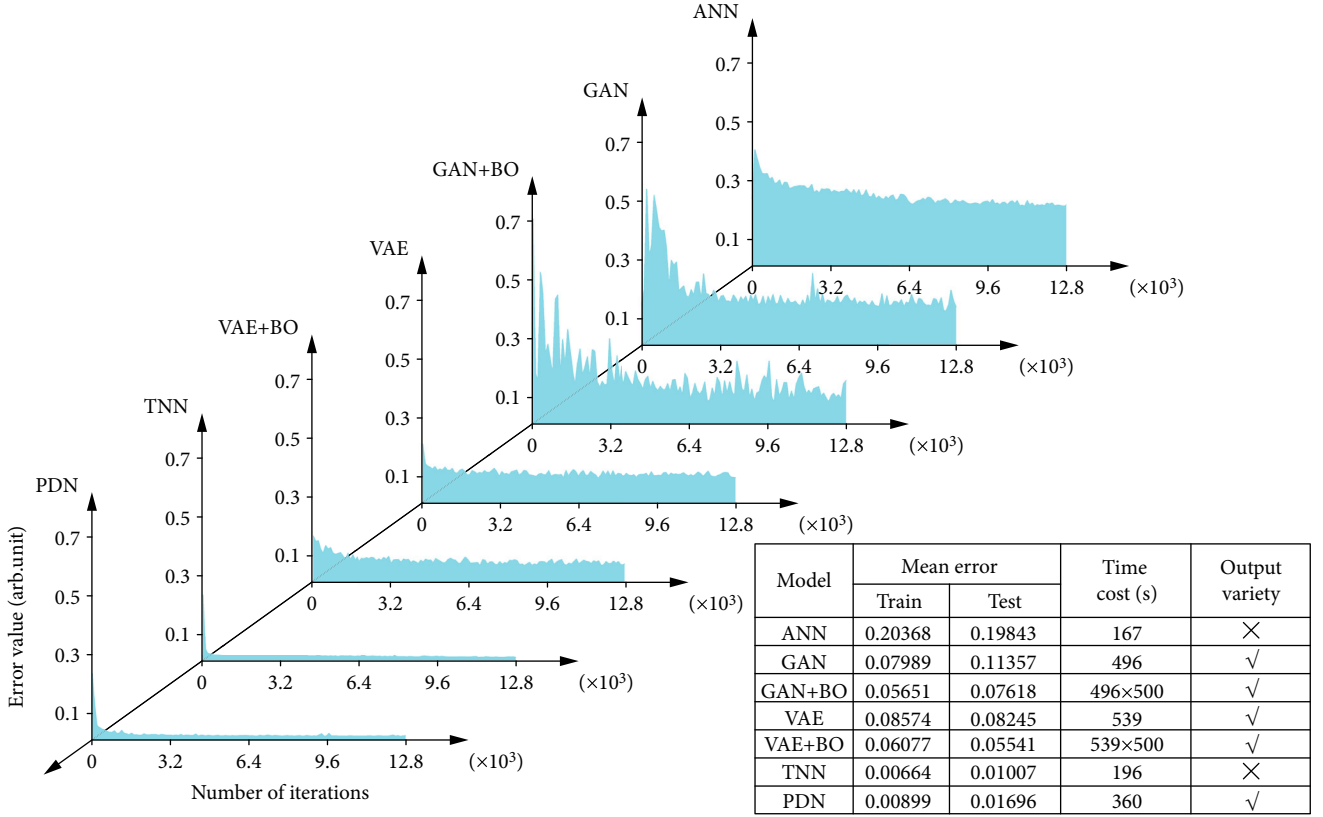

FIGURE 4: Comparison of different deep learning models for inverse design. In comparison, mean error characterizes the accuracy of a model, where the test error reflects the model's generalization ability. The convergence time and output variety reflect the model's optimization efficiency and the varied outputs for the need of multivalued design, respectively. From the table, we find that the ANN and TNN fail to optimize for multivalued function, although the TNN leads to very good accuracy. The GAN and VAE have output varieties but face the trade-off between the accuracy and time costs, whether Bayesian optimization is deployed or not. Only the PDN leads to a balanced performance in all three qualities.

Supplementary Materials and Figs. S3–S5. It is further shown that the PDN can capture slight differences with several similar targets, indicating that the PDN fully considers the effects of uncertainty.

*2.4. Future Vision of Adaptive Inverse Design.* Besides the output variety introduced by the PDN, there are many other approaches that may extend this work to be even more flexible and adaptive in terms of inverse design. For example, the current output size of the PDN is still fixed to be a five-layer geometry, but a module like recurrent neural network or long-short-term memory can be added to the end of the PDN to make predictions absolutely flexible. Likewise, an acoustic embedding that can embed different sizes of transmission spectra or other functionalities can make the input size adaptive. For example, we can take either transmission or reflection with 250 or 50 sampled frequencies to get multiple results that can be five layers or eight layers. This is technically doable and can help the PDN improve design quality and avoid bad designs.

Moreover, since the PDN provides a complete data modeling for the inverse relation in a metastructure, we can leverage this to help understand the physics in metamaterials. Nowadays, data-driven physical equation discovery has succeeded in many chaotic systems, such as turbulence problem [42], and this direction has attracted core attention not only in computational physics but also in artificial intelligence area. It has been shown that not only the known complicated equations like the Navier-Stokes equation but also some unknown or unconfirmed simultaneous equations can be robustly found by the current artificial intelligence algorithms. An analytical physical equation to describe inverse design can be adaptive to the entire universe, which makes it training-free and elegant. The PDN as a complete modeling can help provide endless data for these data-driven algorithms to discover universal equations to solve inverse design in the near future.

## 3. Discussion

In summary, we have demonstrated a probability-density-based deep learning approach, i.e., PDN architecture, which can solve the multivalued inverse-design problems for implementing physically realizable target functionalities in metastructures with high efficiency and accuracy. In acoustics, but not confined to this field, we have successfully employed the PDN to evaluate all the plausible metastructures for different target transmission spectra. The output of the PDN is a superposed probability density, for which the amplitude characterizes how closely the metastructure fits the desired functionalities. This superposition differs on the PDN from baseline ANN that simply has multiple outputs. As a universal mathematical model, the PDN can also be adaptive for many other applied sciences that involve high-dimensional optimization and inverse design. To verify the predictions from the PDN, we design metastructures corresponding to the local maxima in probability density distributions for experimental demonstrations. The measured transmission spectra agree well with both the target and the predicted ones from PDN models. Our results show the superiority of the PDN over other models, with a balanced performance in accuracy, efficiency, and output variety. The proposed PDN is scalable and unparalleled in the aspect of multi-valued inverse design, which paves the way for fast design of novel devices with complex functionalities by optimizing the geometries and/or other parameters and providing a complete modeling to help understand physics in metamaterials.

## 4. Materials and Methods

*4.1. Simulation for Data Preparation.* Data preparation for the transmissions of all labelled structures is relying on state-of-the-art high-precision numerical methods that solve a series of partial differential equations with the consideration of inevitable viscothermal damping. With the support of COMSOL Multiphysics 5.3™, a powerful commercial simulation software, we built up the simulation base model of acoustic metastructures that consists of tunable structural parameters, *i.e.*, the radii of air channels, and regulated boundary conditions and dimensions. Using the interface of MATLAB, we modified the tunable parameters in each simulation cycle to obtain transmission spectra of different metastructures. To speed up the process of for-loop simulations, we distributed cloud servers and collected the transmission-structure paired dataset. The base model and MATLAB codes are provided in the following link: https://github.com/yingtaoluo/Probabilistic-density-network.

*4.2. Acoustic Experiments.* In experiments, the transmission spectra were measured in a commercial impedance tube (Brüel and Kjær type-4206T) with a diameter of 29 mm, shown in Supplementary Materials. The digital signal generated in the computer was sent to the power amplifier and then driving the loudspeaker. Four 1/4-inch condenser microphones (Brüel & Kjær type-4187) are located at different positions to extract the amplitude and phase of local pressure. Before measuring the sample, we measured the transmission of a hollow pipe to finish the calibration and verified the sealing of sound waves. We utilized the standard pitch at 1000 Hz to calibrate the four microphones, made multiple measurements by switching the microphone positions, and obtained the transmission spectrum via the transfer matrix method. In the measurement, the fabricated sample was embedded into the impedance tube with fastening rings to prevent the sound leakage.

*4.3. Neural Network Training.* To train the deep learning models mentioned in this work, we used nearly the same hyperparameters to ensure the fairness. For ANN, GAN, VAE, TNN, and PDN models in Figure 4, we used the same architecture by setting the input dimension of 250 and mapping them into the first hidden layer of 400 neurons. The second, third, and fourth layers have 800, 1600, and 3200 neurons, respectively. Finally, at the output layer, we set 5 neurons. Therefore, all the 5 models could have nearly the same fitting capacity. Moreover, we set the same learning rate of $1 \times 10^{-4}$, the batch size of 256, the epoch of 1000, and the power of regularization of 0. Here, the optimization algorithm is chosen to be Adam, and the activation function is the truncated rectified linear unit. For each individual model, we set their unique hyperparameters without careful selections. For example, for the VAE and GAN, the latent dimension is 10, with the condition added by direct concatenation in the input layer; for the PDN alone, the number of Gaussian distributions is set to be 50. We use batch normalization to avoid overfitting and accelerate training. For all models, we use the same computing device of Tesla P100 for training. The fitting model and the acquisition function used in Bayesian optimization are Gaussian process and expected improvement, and we iterate 500 times to find the optimal hyperparameters for VAE/GAN.

*4.4. Model Generalization Ability.* Generalization is an indicator describing how well a deep learning model fits to the unbiased realistic data rather than the biased training data. A deep learning model can naturally perform very well on the training data but not necessarily well on the test data, as the test data can locate in the interpolated zone where the training data does not cover. In this work, we randomly split the training and test sets without making any intersections between them. Therefore, the values of the test error in Figure 4 reflect both accuracy and generalization ability of the deep learning models.

## Conflicts of Interest

The authors declare no competing interests.

## Authors' Contributions

Y.-T. L., P.-Q. L., and D.-T. L. equally contributed to this work. X.-F. Z., Y.-T. L., and Y.-G. P. conceived the idea. Y.-G. P. prepared the training data, and Y.-T. L developed the PDN architecture. P.-Q. L. designed and fabricated the samples. D.-T. L., P.-Q. L., and S.-H. X. performed the experiments. Y.-T. L. and X.-F. Z. prepared the manuscript. All authors contributed in the data analysis, discussion, and revision of the manuscript. X.-F. Z., Y. L., A. A., and J. Z. supervised this project.

## Acknowledgments

This work is financially supported by the National Natural Science Foundation of China (Grant Nos. 11674119, 11704284, 11774297, 11690030, and 11690032). J. Z. acknowledges the

financial support from the General Research Fund of Hong Kong Research Grants Council (Grant No. 15205219). Y. G. P. and A. A. acknowledge the support of the National Science Foundation and the Simons Foundation. X.-F. Z. acknowledges the Bird Nest Plan of HUST.

## Supplementary Materials

Section S1: statistical theorem of probability-density-based networks. Section S2: technical details of the PDN architecture. Section S3: principal component analysis. Section S4: acoustic experiments on transmission measurement. Fig. S1: neural architecture of PDN. Fig. S2: additional examples for PDN-based reverse designs. Fig. S3: robustness of PDN towards similar target functionalities. Fig. S4: PDN versus direction interpolation. Fig. S5: acoustic functionalities of interpolated structures. Fig. S6: visualization of the principal component analysis. Fig. S7: acoustic standing wave tube. *(Supplementary Materials)*

## References

[1] T. Ergin, N. Stenger, P. Brenner, J. B. Pendry, and M. Wegener, "Three-dimensional invisibility cloak at optical wavelengths," *Science*, vol. 328, no. 5976, pp. 337–339, 2010.

[2] X. Zhu, B. Liang, W. Kan, X. Zou, and J. Cheng, "Acoustic cloaking by a superlens with single-negative materials," *Physical Review Letters*, vol. 106, no. 1, article 014301, 2011.

[3] J. Valentine, J. Li, T. Zentgraf, G. Bartal, and X. Zhang, "An optical cloak made of dielectrics," *Nature Materials*, vol. 8, no. 7, pp. 568–571, 2009.

[4] J. B. Pendry, "Negative refraction makes a perfect lens," *Physical Review Letters*, vol. 85, no. 18, pp. 3966–3969, 2000.

[5] N. Fang, H. Lee, C. Sun, and X. Zhang, "Sub–diffraction-limited optical imaging with a silver superlens," *Science*, vol. 308, no. 5721, pp. 534–537, 2005.

[6] J. Zhu, J. Christensen, J. Jung et al., "A holey-structured metamaterial for acoustic deep-subwavelength imaging," *Nature Physics*, vol. 7, no. 1, pp. 52–55, 2011.

[7] Z. Liu, H. Lee, Y. Xiong, C. Sun, and X. Zhang, "Far-field optical hyperlens magnifying sub-diffraction-limited objects," *Science*, vol. 315, no. 5819, pp. 1686–1686, 2007.

[8] B. Edwards, A. Alù, M. E. Young, M. Silveirinha, and N. Engheta, "Experimental verification of epsilon-near-zero metamaterial coupling and energy squeezing using a microwave waveguide," *Physical Review Letters*, vol. 100, no. 3, article 033903, 2008.

[9] R. Liu, Q. Cheng, T. Hand et al., "Experimental demonstration of electromagnetic tunneling through an epsilon-near-zero metamaterial at microwave frequencies," *Physical Review Letters*, vol. 100, no. 2, article 023903, 2008.

[10] X. Huang, Y. Lai, Z. H. Hang, H. Zheng, and C. T. Chan, "Dirac cones induced by accidental degeneracy in photonic crystals and zero-refractive-index materials," *Nature Materials*, vol. 10, no. 8, pp. 582–586, 2011.

[11] Y. Xie, W. Wang, H. Chen, A. Konneker, B. I. Popa, and S. A. Cummer, "Wavefront modulation and subwavelength diffractive acoustics with an acoustic metasurface," *Nature Communications*, vol. 5, no. 1, p. 5553, 2014.

[12] N. Yu and F. Capasso, "Flat optics with designer metasurfaces," *Nature Materials*, vol. 13, no. 2, pp. 139–150, 2014.

[13] W. Kan, V. M. García-Chocano, F. Cervera et al., "Broadband acoustic cloaking within an arbitrary hard cavity," *Physical Review Applied*, vol. 3, no. 6, article 064019, 2015.

[14] W. X. Jiang, C. W. Qiu, T. C. Han et al., "Broadband all-dielectric magnifying lens for far-field high-resolution imaging," *Advanced Materials*, vol. 25, no. 48, pp. 6963–6968, 2013.

[15] Y. Zhu, J. Hu, X. Fan et al., "Fine manipulation of sound via lossy metamaterials with independent and arbitrary reflection amplitude and phase," *Nature Communications*, vol. 9, no. 1, p. 1632, 2018.

[16] Y. Zhu and B. Assouar, "Systematic design of multiplexed-acoustic-metasurface hologram with simultaneous amplitude and phase modulations," *Physical Review Materials*, vol. 3, no. 4, article 045201, 2019.

[17] K. Melde, A. G. Mark, T. Qiu, and P. Fischer, "Holograms for acoustics," *Nature*, vol. 537, no. 7621, pp. 518–522, 2016.

[18] L. Zigoneanu, B.-I. Popa, and S. A. Cummer, "Three-dimensional broadband omnidirectional acoustic ground cloak," *Nature Materials*, vol. 13, no. 4, pp. 352–355, 2014.

[19] Z. Liu, X. Zhang, Y. Mao et al., "Locally resonant sonic materials," *Science*, vol. 289, no. 5485, pp. 1734–1736, 2000.

[20] G. Ma and P. Sheng, "Acoustic metamaterials: from local resonances to broad horizons," *Science Advances*, vol. 2, no. 2, article e1501595, 2016.

[21] Y. Jin, B. Bonello, R. P. Moiseyenko, Y. Pennec, O. Boyko, and B. Djafari-Rouhani, "Pillar-type acoustic metasurface," *Physical Review B*, vol. 96, no. 10, article 104311, 2017.

[22] Q. Ma, G. D. Bai, H. B. Jing, C. Yang, L. Li, and T. J. Cui, "Smart metasurface with self-adaptively reprogrammable functions," *Light: Science & Applications*, vol. 8, no. 1, p. 98, 2019.

[23] Q. Zhang, C. Liu, X. Wan et al., "Machine-learning designs of anisotropic digital coding metasurfaces," *Advanced Theory and Simulations*, vol. 2, no. 2, article 1800132, 2019.

[24] R. L. Johnston, "Evolving better nanoparticles: genetic algorithms for optimising cluster geometries," *Dalton Transactions*, vol. 32, no. 22, pp. 4193–4207, 2003.

[25] M. B. Giles and N. A. Pierce, "An introduction to the adjoint approach to design," *Flow, Turbulence and Combustion*, vol. 65, no. 3/4, pp. 393–415, 2000.

[26] Y. E. Lee, O. D. Miller, M. T. Homer Reid, S. G. Johnson, and N. X. Fang, "Computational inverse design of non-intuitive illumination patterns to maximize optical force or torque," *Optics Express*, vol. 25, no. 6, pp. 6757–6766, 2017.

[27] Y. LeCun, Y. Bengio, and G. Hinton, "Deep learning," *Nature*, vol. 521, no. 7553, pp. 436–444, 2015.

[28] M. H. S. Segler, M. Preuss, and M. P. Waller, "Planning chemical syntheses with deep neural networks and symbolic AI," *Nature*, vol. 555, no. 7698, pp. 604–610, 2018.

[29] G. Carleo and M. Troyer, "Solving the quantum many-body problem with artificial neural networks," *Science*, vol. 355, no. 6325, pp. 602–606, 2017.

[30] J. Behler and M. Parrinello, "Generalized neural-network representation of high-dimensional potential-energy surfaces," *Physical Review Letters*, vol. 98, no. 14, p. 146401, 2007.

[31] S. Molesky, Z. Lin, A. Y. Piggott, W. Jin, J. Vucković, and A. W. Rodriguez, "Inverse design in nanophotonics," *Nature Photonics*, vol. 12, no. 11, pp. 659–670, 2018.

[32] W. Ma, F. Cheng, Y. Xu, Q. Wen, and Y. Liu, "Probabilistic representation and inverse design of metamaterials based on a deep generative model with semi-supervised learning strategy," *Advanced Materials*, vol. 31, no. 35, p. 1901111, 2019.

[33] G. E. Hinton, "Deterministic Boltzmann learning performs steepest descent in weight-space," *Neural Computation*, vol. 1, no. 1, pp. 143–150, 1989.

[34] M. Arjovsky, S. Chintala, and L. Bottou, "Wasserstein Generative Adversarial Networks," in *Proceedings of the Thirty-fourth International Conference on Machine Learning*, pp. 214–223, Sydney, Australia, 2017.

[35] K. Roth, A. Lucchi, S. Nowozin, and T. Hofmann, "Stabilizing Training of Generative Adversarial Networks through Regularization," in *Proceedings of the 30th Advances in Neural Information Processing Systems*, pp. 2018–2028, Long Beach, CA, USA, 2017.

[36] I. Gulrajani, F. Ahmed, M. Arjovsky, V. Dumoulin, and A. C. Courville, "Improved Training of Wasserstein GANs," in *Proceedings of the 30th Advances in Neural Information Processing Systems*, pp. 5767–5777, Long Beach, CA, USA, 2017.

[37] J. Bergstra and Y. Bengio, "Random search for hyper-parameter optimization," *Journal of Machine Learning Research*, vol. 13, pp. 281–305, 2012.

[38] D. Liu, Y. Tan, E. Khoram, and Z. Yu, "Training deep neural networks for the inverse design of nanophotonic structures," *ACS Photonics*, vol. 5, no. 4, pp. 1365–1369, 2018.

[39] P. Baldi, "Autoencoders, Unsupervised Learning, and Deep Architectures," in *Proceedings of ICML Workshop on Unsupervised and Transfer Learning*, pp. 37–49, Edinburgh, Scotland, 2012.

[40] D. P. Kingma and M. Welling, "Auto-encoding variational bayes," 2013, https://arxiv.org/abs/1312.6114.

[41] S. Wold, K. Esbensen, and P. Geladi, "Principal component analysis," *Chemometrics and Intelligent Laboratory Systems*, vol. 2, no. 1-3, pp. 37–52, 1987.

[42] S. H. Rudy, S. L. Brunton, J. L. Proctor, and J. N. Kutz, "Data-driven discovery of partial differential equations," *Science Advances*, vol. 3, no. 4, article e1602614, 2017.